%% file: root.tex
\documentclass[conference]{IEEEtran}
\IEEEoverridecommandlockouts
\usepackage{graphics} 
\usepackage{epsfig} 
\usepackage{mathptmx} 
\usepackage{times} 
\usepackage{microtype}
\usepackage{algorithmic}
\usepackage{textcomp}
\usepackage{xcolor}
\usepackage{gensymb}
\usepackage{caption}
\usepackage{hyperref}
\usepackage{mwe}
\usepackage{verbatim}
\usepackage{amsbsy}
\usepackage{siunitx}
\usepackage{tabularx,booktabs}
\usepackage{dblfloatfix}
\usepackage{cite}
\usepackage{bm}
\usepackage{amsmath,amssymb,amsfonts}
\usepackage{subcaption}
\usepackage{float}

\usepackage{algorithm, algorithmic}
\usepackage[autolanguage]{numprint}

\usepackage{spreadtab}
\usepackage{multirow}
\usepackage{cite}
\usepackage{hyperref}
\usepackage{xcolor}
\usepackage{algorithmic}
\usepackage{textcomp}
\usepackage{xcolor}
\usepackage{gensymb}
\usepackage{caption}
\usepackage{mwe}
\usepackage{amsfonts}
\usepackage{amssymb}
\usepackage{verbatim}
\usepackage{amsbsy}
\usepackage{siunitx}
\usepackage{tabularx, booktabs}
\usepackage{soul}
\usepackage{tikz}
\usetikzlibrary{calc}
\def\BibTeX{{\rm B\kern-.05em{\sc i\kern-.025em b}\kern-.08em
    T\kern-.1667em\lower.7ex\hbox{E}\kern-.125emX}}

\makeatletter
\newif\if@anonymize

\@anonymizetrue    

\if@anonymize
  \newcommand{\highlight@DoHighlight}{
    \fill [outer sep = -15pt, inner sep = 0pt, color=black]
          ($(begin highlight)+(0,8pt)$) rectangle ($(end highlight)+(0,-3pt)$) ;
  }

  \newcommand{\highlight@BeginHighlight}{
    \coordinate (begin highlight) at (0,0) ;
  }

  \newcommand{\highlight@EndHighlight}{
    \coordinate (end highlight) at (0,0) ;
  }

  \newdimen\highlight@previous
  \newdimen\highlight@current
  \newlength{\item@width}

  \DeclareRobustCommand*\anonymize{%
    \SOUL@setup
    \def\SOUL@preamble{%
      \begin{tikzpicture}[overlay, remember picture]
        \highlight@BeginHighlight
        \highlight@EndHighlight
      \end{tikzpicture}%
    }%
    \def\SOUL@postamble{%
      \begin{tikzpicture}[overlay, remember picture]
        \highlight@EndHighlight
        \highlight@DoHighlight
      \end{tikzpicture}%
    }%
    \def\SOUL@everyhyphen{%
      \discretionary{%
        \SOUL@setkern\SOUL@hyphkern
        \SOUL@sethyphenchar
        \tikz[overlay, remember picture] \highlight@EndHighlight ;%
      }{%
      }{%
        \SOUL@setkern\SOUL@charkern
      }%
    }%
    \def\SOUL@everyexhyphen##1{%
      \SOUL@setkern\SOUL@hyphkern
      \settowidth{\item@width}{##1}%
      \makebox[\item@width]{}%
      \discretionary{%
        \tikz[overlay, remember picture] \highlight@EndHighlight ;%
      }{%
      }{%
        \SOUL@setkern\SOUL@charkern
      }%
    }%
    \def\SOUL@everysyllable{%
      \begin{tikzpicture}[overlay, remember picture]
        \path let \p0 = (begin highlight), \p1 = (0,0) in \pgfextra
          \global\highlight@previous=\y0
          \global\highlight@current =\y1
        \endpgfextra (0,0) ;
        \ifdim\highlight@current < \highlight@previous
          \highlight@DoHighlight
          \highlight@BeginHighlight
        \fi
      \end{tikzpicture}%
      \settowidth{\item@width}{\the\SOUL@syllable}%
      \makebox[\item@width]{}%
      \tikz[overlay, remember picture] \highlight@EndHighlight ;%
    }%
    \SOUL@
  }
\else
  \newcommand{\anonymize}[1]{#1}
\fi

\newcommand{\linebreakand}{%
  \end{@IEEEauthorhalign}
  \hfill\mbox{}\par
  \mbox{}\hfill\begin{@IEEEauthorhalign}
}

\makeatother

\begin{document}

\title{Parallel Distributional Prioritized Deep Reinforcement Learning for Unmanned Aerial Vehicles\\
}

\author{\IEEEauthorblockN{1\textsuperscript{st} Alisson Henrique Kolling}
\IEEEauthorblockA{\textit{Center for Computational Science} \\
\textit{Federal University of Rio Grande}\\
Rio Grande, Brazil\\
alikolling@gmail.com}
\and
\IEEEauthorblockN{2\textsuperscript{nd} Victor Augusto Kich}
\IEEEauthorblockA{\textit{Intelligent Robot Laboratory}\\
\textit{University of Tsukuba}\\
Tsukuba, Japan \\
victorkich98@gmail.com}
\and
\IEEEauthorblockN{3\textsuperscript{rd} Junior Costa de Jesus}
\IEEEauthorblockA{\textit{VersusAI Team} \\
Sapezal, Brazil\\
dranaju@gmail.com}
\and
\IEEEauthorblockN{4\textsuperscript{th} Andressa Cavalcante da Silva}
\IEEEauthorblockA{\textit{Center for Computational Science} \\
\textit{Federal University of Rio Grande}\\
Rio Grande, Brazil\\
andressacavalcante94@gmail.com}
\and
\IEEEauthorblockN{5\textsuperscript{th} Ricardo Bedin Grando}
\IEEEauthorblockA{\textit{Robotics and Artificial Inteligence Lab} \\
\textit{Technological University of Uruguay}\\
Rivera, Uruguay \\
0000-0002-2939-5304}
\and
\IEEEauthorblockN{6\textsuperscript{th} Paulo Lilles Jorge Drews-Jr}
\IEEEauthorblockA{\textit{Center for Computational Science} \\
\textit{Federal University of Rio Grande}\\
Rio Grande, Brazil \\
paulodrews@furg.br
}
\linebreakand
\IEEEauthorblockN{7\textsuperscript{th} Daniel F. T. Gamarra}
\IEEEauthorblockA{\textit{DPEE}\\
\textit{Federal University of Santa Maria}\\
Santa Maria, Brazil \\
fernandotg99@gmail.com}
}

\maketitle

\thispagestyle{empty}
\pagestyle{empty}

\input{sections/1_abstract}
\input{sections/2_introduction}
\input{sections/3_related_works}
\input{sections/4_theorethical}
\input{sections/5_experimental}
\input{sections/7_results}
\input{sections/8_conclusion}
\input{sections/9_aknowledgement}

\input{sections/10_references}

\end{document}

%% file: sections/1_abstract.tex
\begin{abstract}
This work presents a study on parallel and distributional deep reinforcement learning applied to the mapless navigation of UAVs. For this, we developed an approach based on the Soft Actor-Critic method, producing a distributed and distributional variant named PDSAC, and compared it with a second one based on the traditional SAC algorithm. In addition, we also embodied a prioritized memory system into them. The UAV used in the study is based on the Hydrone vehicle, a hybrid quadrotor operating solely in the air. The inputs for the system are 23 range findings from a Lidar sensor and the distance and angles towards a desired goal, while the outputs consist of the linear, angular, and, altitude velocities. The methods were trained in environments of varying complexity, from obstacle-free environments to environments with multiple obstacles in three dimensions.
The results obtained, demonstrate a concise improvement in the navigation capabilities by the proposed approach when compared to the agent based on the SAC for the same amount of training steps.
In summary, this work presented a study on deep reinforcement learning applied to mapless navigation of drones in three dimensions, with promising results and potential applications in various contexts related to robotics and autonomous air navigation with distributed and distributional variants.

\end{abstract}

%% file: sections/2_introduction.tex
\section{Introduction}\label{introduction}

Unmanned aerial vehicles (UAVs), commonly called drones, are already found significantly in everyday applications in the civil area, commercial, and military areas \cite{kitonsa2018significance}. This is because it does not have a human operator or pilot in person, enabling the operation of these systems economically in previously unfeasible and dangerous locations and conditions. With the increase in the use of these unmanned systems, some challenges arise, mainly in developing the level of autonomy of these systems.

The growth in the number of autonomous systems requires a series of developments in artificial intelligence and decision-making. One of the methods that stand out in solving this problem is Deep Reinforcement Learning (DRL). DRL provides a learning framework that enables agents to perform optimally through sequential interactions with their environment. By incorporating deep neural networks, it can defeat human experts playing various Atari \cite{mnih_playing_2013}, Starcraft II \cite{starcraft} and Dota-2 \cite{dota2} video games, proving the effectiveness of such methods and showing the possibility of their application in the most diverse areas that require the control of an agent.
These features of DRL have led to many research efforts on their application in autonomous vehicles. DRL has been mainly applied to control system tasks that regulate the speed, attitude, and navigation of many UAVs~\cite{ricardo2021icra,dranaju_iros,grando_lars_22,walker2019deep, 9307024}.

Overall, these methods have great potential for application for autonomous aerial vehicles, as they are used in complex environments where classic methods exhibit difficulties. However, DRL methods need some time to train their neural networks and reach their maximum performance and tend to present problems related to catastrophic forgetting given the complexity that some neural networks may present~\cite{cahill2011catastrophic, pfülb2019comprehensive}. Thus, wishing to speed up the training and present more stable DRL agents, a parallel version of the Distributional Soft Actor-Critic (DSAC) method - named Parallel Distributional Soft Actor-Critic (PDSAC) - is presented in this work. We compare with the method Soft Actor-Critic (SAC) to exemplify those improvements. With this, this work shows the following contributions as results: 
\begin{itemize}
    \item We develop a new distributed distributional DRL method, named PDSAC.
    \item We managed to reduce the training time of DRL agents by utilizing parallel agents during training.
    \item We aim to improve the robustness of the agents by employing distributional critics.
\end{itemize}


%% file: sections/3_related_works.tex
\section{Related Works}
\label{related_works}

The area of DRL has received some attention recently due to the progress and the great results obtained when applying DRL techniques to problems in Robotics. Some examples of this advance would be the results obtained in electronic games and board games, where techniques have decisively overcome human beings, as demonstrated by Mnih \emph{et al.} in \cite{Mnih2015}. Another example would be the use in controlling robots, with Jesus \emph{et al.}  in \cite{deJesus2021} demonstrating similar and superior results to those of classical techniques.

In the application of DRL to UAVs, some works can be highlighted. The work of Rodriguez \emph{et al.} \cite{rodriguez-ramos_deep_2019} is highlighted, who employed a DRL algorithm called \textit{Deep Deterministic Policy Gradient} (DDPG) to allow an advanced autonomous drone to maneuver and land on a mobile platform. The authors integrate the DDPG algorithm into their reinforcement learning simulation framework implemented using Gazebo and \textit{Robot Operating System} (ROS). Another work that has a lot of influence on this one is that of Grando \emph{et al.} \cite{ricardo2021icra}, where he evaluates DRL methods in aerial and aquatic navigation of a hybrid drone. The employed methods consist of stochastic and deterministic methods. The hybrid drone simulated model used in our work is the same used by Grando \emph{et al.}, but only for aerial navigation.

Deep-RL or DRL techniques are traditionally based on single-value expectations. However, Bellemare \emph{et al.}~\cite{bellemare_c51} introduced a distributional approach that models return expectations as probabilistic distributions. Duan \emph{et al.}~\cite{dsac} extended this idea to the SAC algorithm~\cite{haarnoja2018soft}, inspiring the approaches in this paper. Distributed Deep-RL accelerates training by distributing computation and data acquisition. Mnih \emph{et al.}~ \cite{Mnih2015} used asynchronous actor-critic methods, while Horgan \emph{et al.}~\cite{horgan2018distributed} proposed Ape-X, which distributed data sampling and extended experience replay. Barth-Maron \emph{et al.}~\cite{barth-maron2018distributional} improved DDPG with Ape-X concepts. 

Overall, this work introduces a novel combination of parallelism and distributional RL techniques, aiming to advance the capabilities of the SAC algorithm in solving complex reinforcement learning problems with continuous action spaces. This work differs from the previous ones by bringing the parallel methodology from Ape-X to the distributional version of SAC. It seeks to enhance the exploration-exploitation trade-off by employing multiple actors in parallel, while also capturing the distributional nature of the action-value function to handle uncertainties and optimize performance.

%% file: sections/4_theorethical.tex
\section{Methodology}
\label{theoretical}

\subsection{Deep Reinforcement Learning}

Reinforcement learning is an area of machine learning that studies the mapping of actions to situations in order to maximize a reward signal. Currently, reinforcement learning methods are employed in the areas of control, robotics, and recommendation systems, and have been very successful in board games and electronic games.

\subsection{SAC}

The Soft Actor-Critic (SAC)~\cite{haarnoja2018soft} is a DRL algorithm that optimizes a stochastic policy in an off-policy manner. The method employs concepts from actor-critic methods, having a network to learn the policy, another to estimate the state-action value function, and a last one to estimate the state-value function. However, it uses the trick of the double critic, having two critic networks compete to find the best estimate of the value function.

The policy is trained with the objective of maximizing the expected return and entropy at the same time:

\begin{equation}
\begin{aligned}
J(\theta) = \sum_{t=1}^T \mathbb{E}{(s_t, a_t) \sim \rho{\pi_\theta}} [r(s_t, a_t) + \alpha \mathcal{H}(\pi_\theta(\cdot \vert s_t))]
\end{aligned}
\end{equation}

where $\mathcal{H}(.)$ is the entropy measure and controls the importance of the entropy term, known as the temperature parameter $\alpha$. Maximizing entropy leads to policies that can explore more and capture multiple modes of near-optimal strategies (i.e., if several options seem to be equally good, the policy should assign each one the same probability of being chosen).


The state-value function is optimized by minimizing the mean squared error between the value estimated by the network and the value calculated using the state-action value function:

\begin{equation}
\begin{aligned}
J_V(\psi) = \mathbb{E}{s_t \sim \mathcal{D}} \Big[&\frac{1}{2} \Big(V\psi(s_t) - \mathbb{E}{a_t \sim \pi\theta(\cdot \vert s_t)}[Q_w(s_t, a_t) - \\
& - \alpha \log \pi_\theta(a_t \vert s_t)] \Big)^2\Big]
\end{aligned}
\end{equation}

The state-action value function aims to minimize the Bellman residual or temporal difference:

\begin{equation}
\begin{aligned}
J_Q(w) =&\ \mathbb{E}{(s_t, a_t) \sim \mathcal{D}} \Big[\frac{1}{2}\Big( Q_w(s_t, a_t) - \Big(r(s_t, a_t) + \\
& + \gamma \mathbb{E}{s_{t+1} \sim \rho_\pi(s)}[V_{\bar{\psi}}(s_{t+1})]\Big) \Big)^2\Big]
\end{aligned}
\end{equation}

where $\mathcal{D}$ is the set of training transitions and $\bar{\psi}$ is the target network for the state-value function.

\subsection{DSAC}

The Distributional Soft Actor-Critic (DSAC)~\cite{dsac} is an extension of the Soft Actor-Critic (SAC) that adds a distributional layer to the value network to learn the distribution of the Q-function instead of its mean value. This allows the algorithm to handle ambiguities in the estimation of the value function, for example, in scenarios with multiple optimal solutions. The goal of DSAC is to maximize the weighted sum of future rewards and policy entropy.

The policy is represented by a neural network that maps a state $s$ to a probability distribution over actions $a$, $\pi_{\theta}(a|s)$. The state-action value function Q is represented by a neural network that maps a state $s$ and action $a$ to a probability distribution over future rewards $r$. The value network uses a set of atoms to represent the probability distribution, instead of a single mean value. Each atom $z_i$ represents a possible future reward, and the distribution is formed by a probability mass $p_i$. The set of atoms is defined a priori, usually using a uniform distribution over a range of values.

The optimization objective is to maximize the expected reward and policy entropy while minimizing the KL divergence between the current distribution of Q and the target distribution of Q. The objective of the value function is to minimize the mean squared error between the current state value and the expected state value.

To update the policy, the gradient can be calculated as:

\begin{equation}
\begin{aligned}
\nabla_{\theta} J(\theta) \approx \frac{1}{B} \sum_{i=1}^B \nabla_{\theta} \log \pi_{\theta}(a_i|s_i) (Q_{\theta}^{s_i, a_i} - V_{\theta}^{s_i})
\end{aligned}
\end{equation}

To update the value function, the gradient can be calculated as:

\begin{equation}
\begin{aligned}
\nabla_{\theta} J_V(\theta) \approx \frac{1}{B} \sum_{i=1}^B \nabla_{\theta} \frac{1}{2} || Q_{\theta}^{s_i, a_i} - T_{\theta}^{s_i, a_i} ||^2
\end{aligned}
\end{equation}

And to update the distribution of Q, the gradient can be calculated as:

\begin{equation}
\begin{aligned}
\nabla_{\theta} J_Q(\theta) \approx \frac{1}{B} \sum_{i=1}^B \nabla_{\theta} D_{KL}(Q_{\theta}^{s_i, a_i} || T_{\theta}^{s_i, a_i})
\end{aligned}
\end{equation}

In summary, the Distributional Soft Actor-Critic algorithm uses the distribution of Q instead of the value of Q to learn the policy. This allows the policy to better explore multiple near-optimal strategies and maximize policy entropy to ensure exploration. Additionally, the algorithm uses a target distribution to update the distribution of Q rather than a target value, which increases the stability of the Q-update.

%% file: sections/5_experimental.tex
\section{Experimental Setup}
\label{experimental_setup}

\subsection{Training setup}

\begin{figure}
     \begin{center}
         \includegraphics[width=0.45\textwidth]{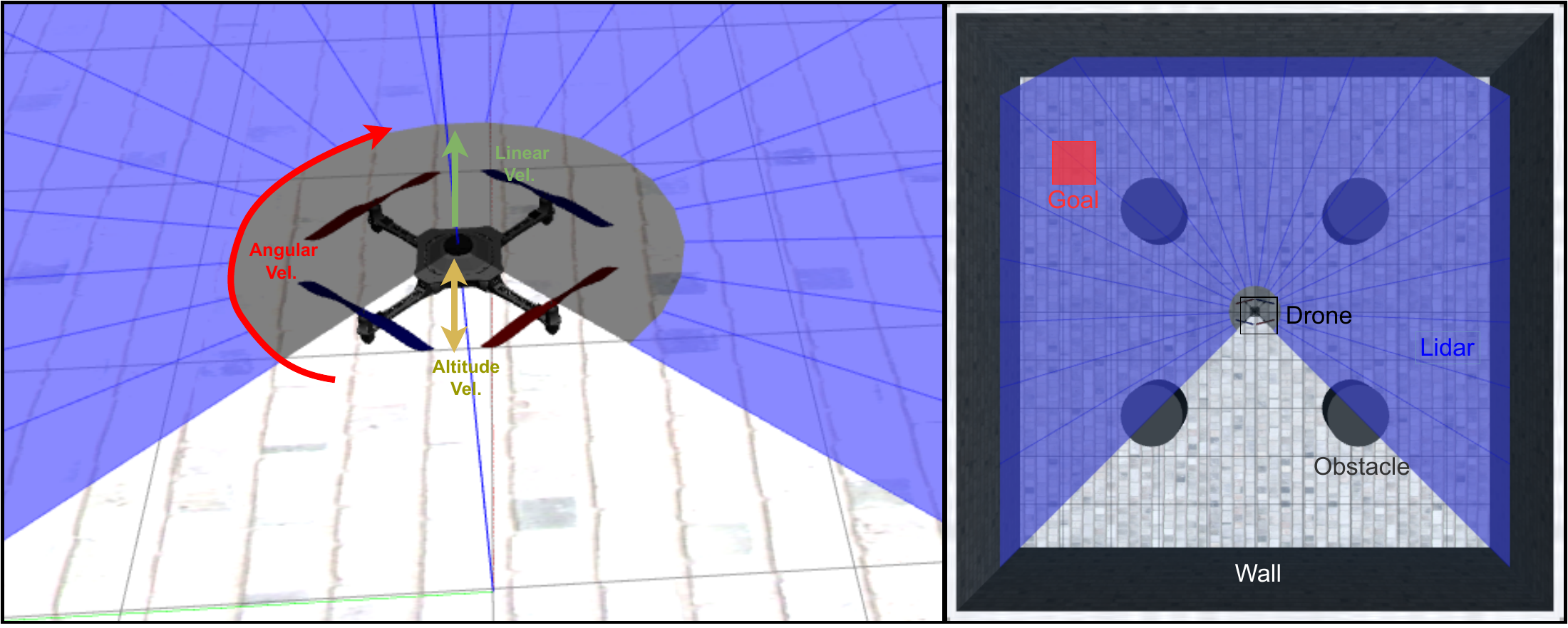}
     \end{center}
     \caption{Drone movement.}
     \label{fig:vels}
     \vspace{-5mm}
\end{figure}

The Robotic Operational System (ROS) acts as an intermediary, managing connections between developed methods, software, and simulated robots. Gazebo and RotorS provide a comprehensive drone simulation environment \cite{furrer_rotorsmodular_2016}, while PyTorch is utilized for developing DRL algorithms. The aerial vehicle used in this work is based on the Hydrone model \cite{drews_hybrid_2014, ricardo2021icra}, controlled by linear and angular velocities in six degrees of freedom $(x, y, z, roll, pitch, yaw)$. In the left Figure \ref{fig:vels} you can see the vehicle presented in the simulation.

\begin{figure}[]
    \begin{center}
 
     \begin{subfigure}{0.15\textwidth}
         \includegraphics[width=\textwidth]{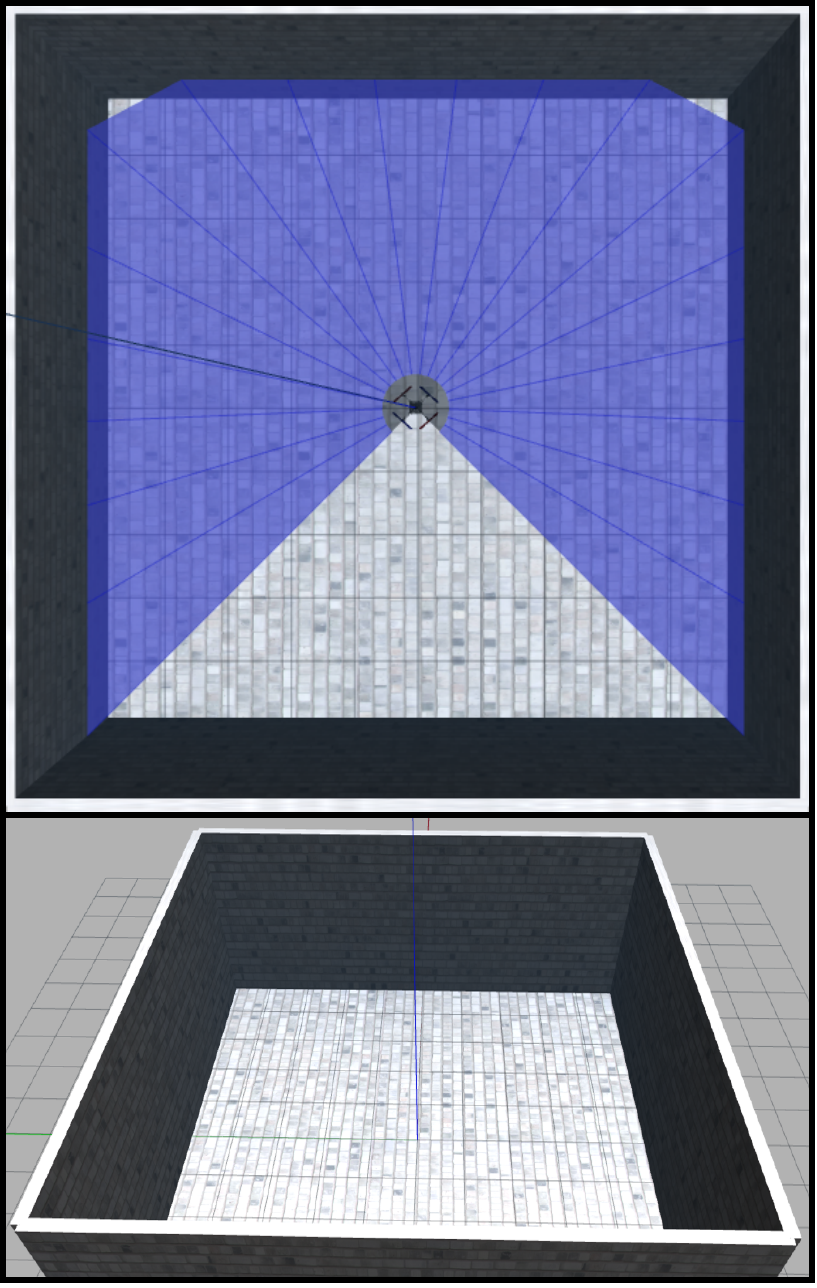}
         \caption{First.}
         \label{subfig:env1}
     \end{subfigure}
     \begin{subfigure}{0.15\textwidth}
         \includegraphics[width=\textwidth]{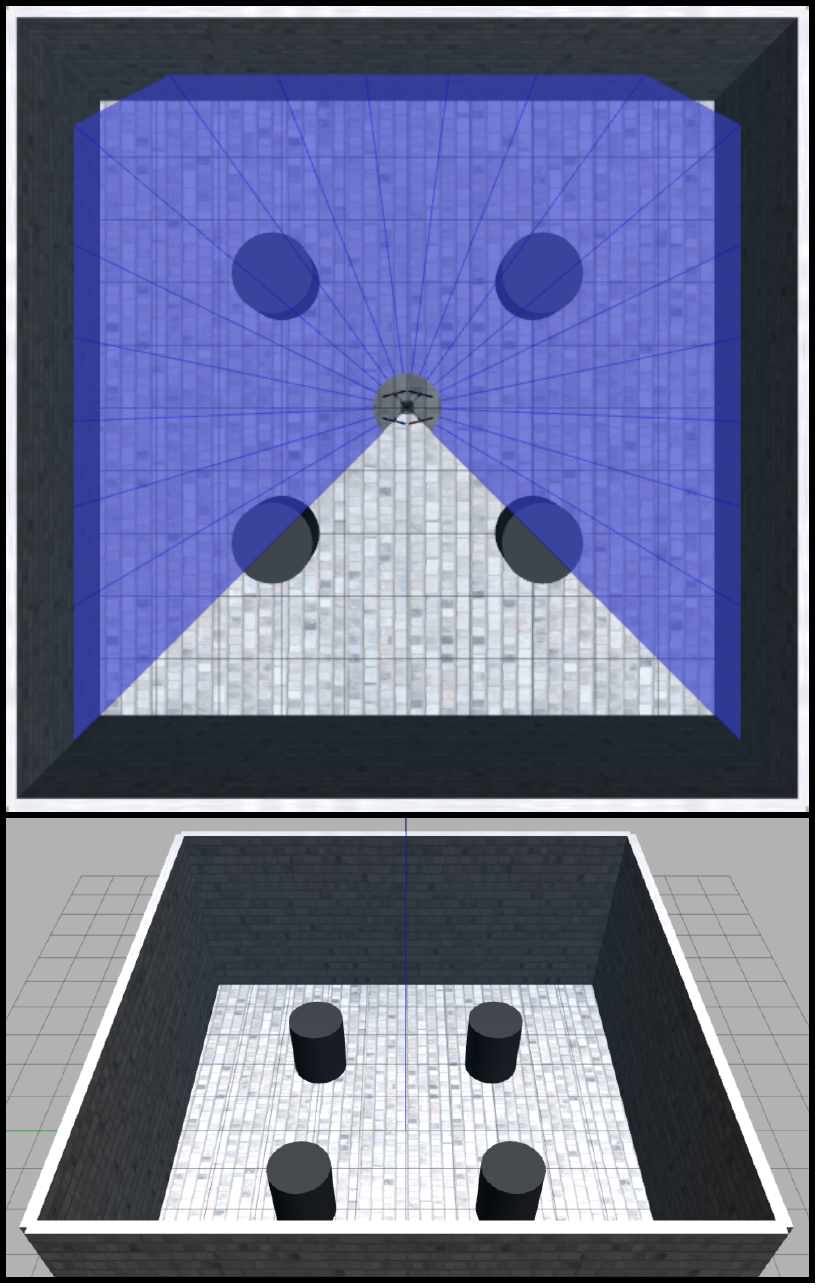}
         \caption{Second.}
         \label{subfig:env2}
     \end{subfigure}
     \begin{subfigure}{0.15\textwidth}
         \includegraphics[width=\textwidth]{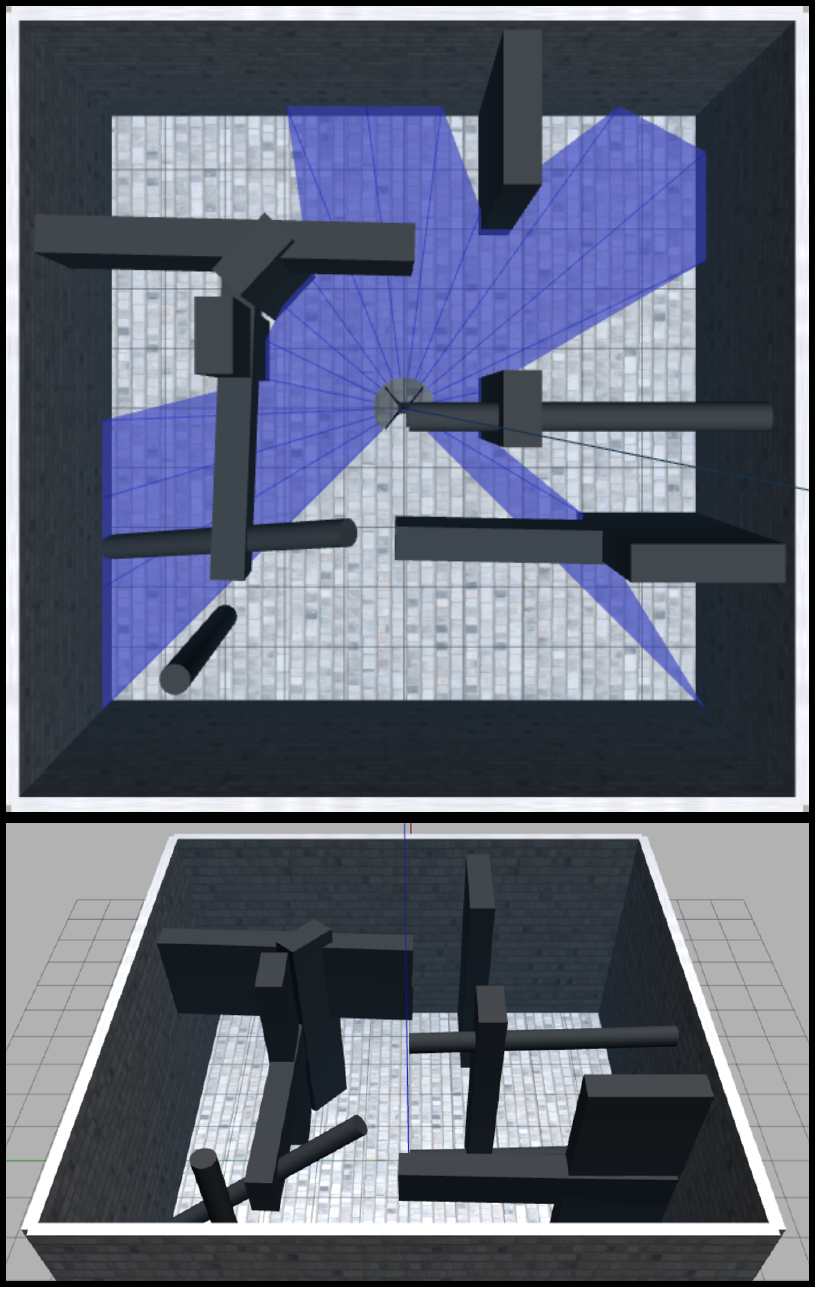}
         \caption{Third.}
         \label{subfig:env3}
     \end{subfigure}
    \end{center}
     \caption{Training environments.}
     \label{fig:env}
     \vspace{-5mm}
\end{figure}

Three training environments were developed, which can be seen in Figure \ref{fig:env}.

In the first environment, Figure \ref{subfig:env1}, the room is empty with only the outer walls as obstacles. The second environment, Figure \ref{subfig:env2}, introduces four additional obstacles inside the room, increasing the challenge for the robot to navigate around them. The last environment, Figure \ref{subfig:env3}, presents a complex navigation scenario with obstacles extending in three dimensions, requiring the robot to consider altitude during navigation. The goal is for the vehicle to autonomously navigate this environment without collisions.

\subsection{Reward Function}



A DRL method relies on a reward function for feedback and improvement. The reward system reinforces actions with high returns and penalizes those with low returns, enabling learning. Developing an effective reward system is empirical and relies on problem-specific knowledge. The implemented reward system is provided below:
\begin{equation}
r(s_t, a_t) =
\begin{cases}
r_{arr}\ \textrm{se} \ d_t < c_d\\
r_{coll}\ \textrm{if}\ min_x < c_o\ \textrm{or}\ min_z < 0.2\ \textrm{or}\ max_z >4.0\\
r_{idle}\ \textrm{se}\ min_x >= c_o\ \textrm{e}\ d_t >= c_d
\end{cases}
\end{equation}



Three types of rewards were used in the experiment. The agent received a reward of $200$ ($r_{arr}$) when it successfully reached the target within a margin of $0.85$ meters ($c_d$). In case of a collision with an obstacle or when reaching the scenario limits, a negative reward of $-20$ ($r_{coll}$) was given. To encourage exploration, an additional reward ($r_{idle}$) was implemented, motivating the robot to move and attempt to reach the goal within the 500 steps. Collisions were detected if the distance sensor readings were below $0.65$ meters ($c_o$), which was determined based on the drone's dimensions. Negative rewards were also assigned for altitudes below $0.2$ meters and above $4.0$ meters.

\subsection{Network Structures}

\begin{figure}[tbp]
     \begin{center}
        \includegraphics[width=0.4\textwidth]{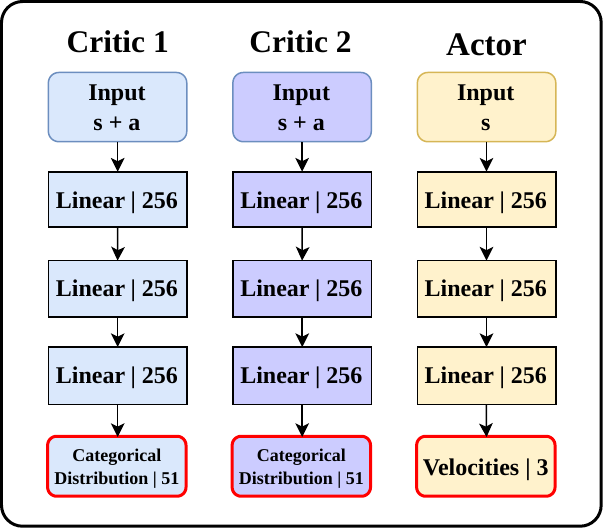}
     \end{center}
     \caption{Neural networks structure.}
     \label{fig:networks}
      \vspace{-5mm}
\end{figure}

The networks of our proposed SAC and PDSAC approaches have three fully connected hidden layers with 256 neurons each, connected through the activation of ReLU. The hyper-parameters in this work were based on previous works on the topic \cite{ricardo2021icra, deJesus2021}. The action varies between $-1$ and $1$, and the hyperbolic tangent function (Tanh) was used as the activation function for the policy network. Outputs scale from $-0.25$ to $0.25$ meters per second for linear velocity, from $-0.1$$m/s$ to $0.1$$m/s$ for angular velocity, and $-0.25 m/s$ and $0.25 m/s$ for altitude change. For both approaches, the Q value of the current state is predicted in the critical network, while the actor-network predicts the action from the current state. Five agents were used in the training phase for parallel approaches: one for evaluation and four for exploration, as can be seen in Figure \ref{fig:networks}.

\subsection{Parallelism}

The difference between the proposed methods is the use of $K$ agents, which are executed simultaneously, thus reducing the time required for training the methods.
PDSAC employs the principles for parallel agents proposed by Dan Horgan \cite{horgan2018distributed}.

Parallel methods work by separating action from learning. For this, there are $K$ processes executing agents that interact with the environments. These agents receive a network copy of the policy through memory queues and are thus able to perform the desired actions. The experiences obtained by the agents are sent to the replay buffer, which serves as a memory for learning. In a separate process, you have neural network learning taking place. This process performs the learning, receiving the experiences from the replay buffer, calculating the errors, updating the networks, and sending the weights of the networks to the agents. To create multiple processes, we used the multiprocessing Python library and its version of PyTorch.
It was also necessary to parallelize the environments so that each agent has its environment. For this purpose, different simulations with ROS and Gazebo were performed on the agents' $K$ processes. This is possible by modifying the TCP/IP address of the master ROS in each process.

\begin{figure}[tbp]
     \begin{center}
         \includegraphics[width=0.4\textwidth]{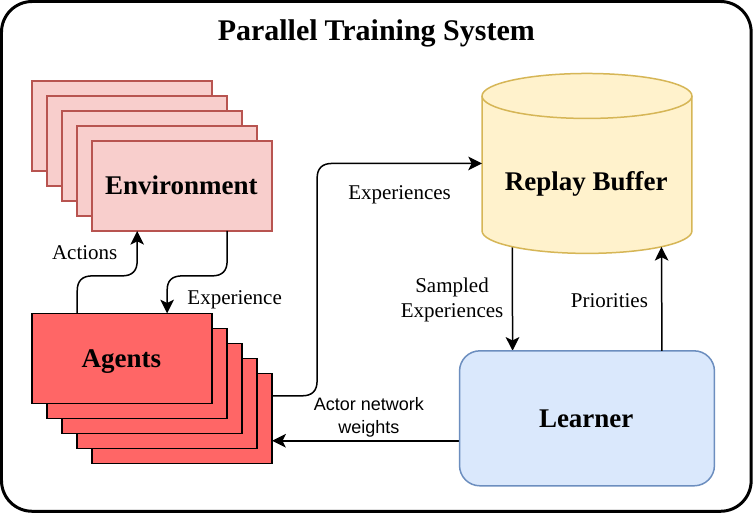}
     \end{center}
     \caption{Parallelism structure.}
     \label{fig:parallel}
     \vspace{-5mm}
\end{figure}

\subsection{Prioritized memory}

The memory of an off-policy DRL method consists of a replay buffer containing the experiences obtained by the agent. Using the experience stored in the memory during the method training, it becomes possible to break the temporal correlations.
One way to improve this memory is to prioritize certain experiences that favor learning \cite{prioritized_memory}. To decide which experiences to prioritize, we use the time difference error, which measures how far the reward value is from your estimate. For this, this algorithm stores the last time difference error found along with each transition in the replay memory.
The transition with the largest absolute time difference error is replayed from memory the most times.
In this work, the prioritized memory model is based on the one developed by Schaul \textit{et al.} \cite{prioritized_memory} called proportional prioritization.

%% file: sections/7_results.tex
\section{Results}
\label{results}

\begin{figure*}[tbp]
\centering
     \begin{subfigure}{0.325\textwidth}
         \includegraphics[width=\textwidth]{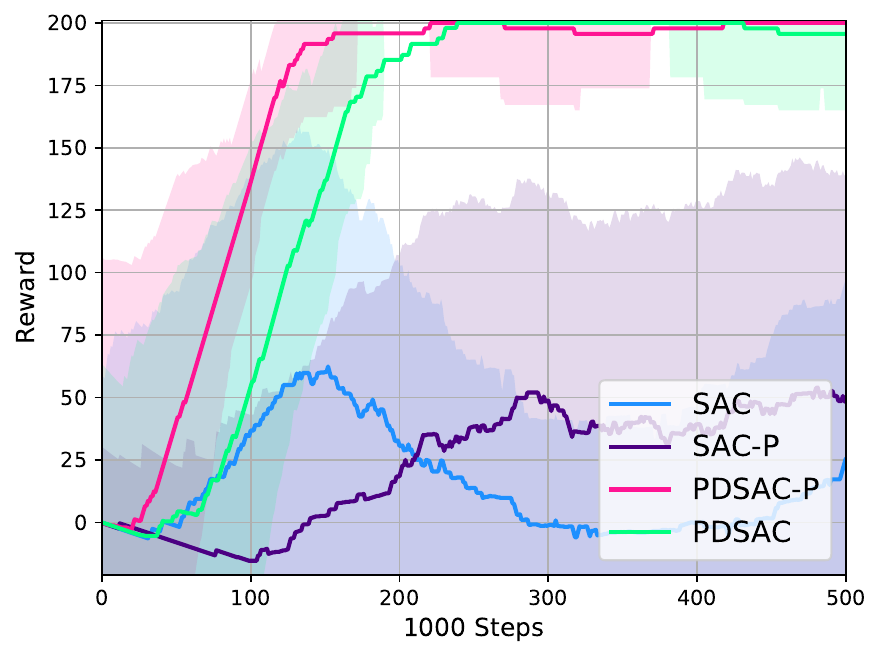}
         \caption{Environment 1.}
         \label{fig:stage1}
     \end{subfigure}
     \begin{subfigure}{0.325\textwidth}
         \includegraphics[width=\textwidth]{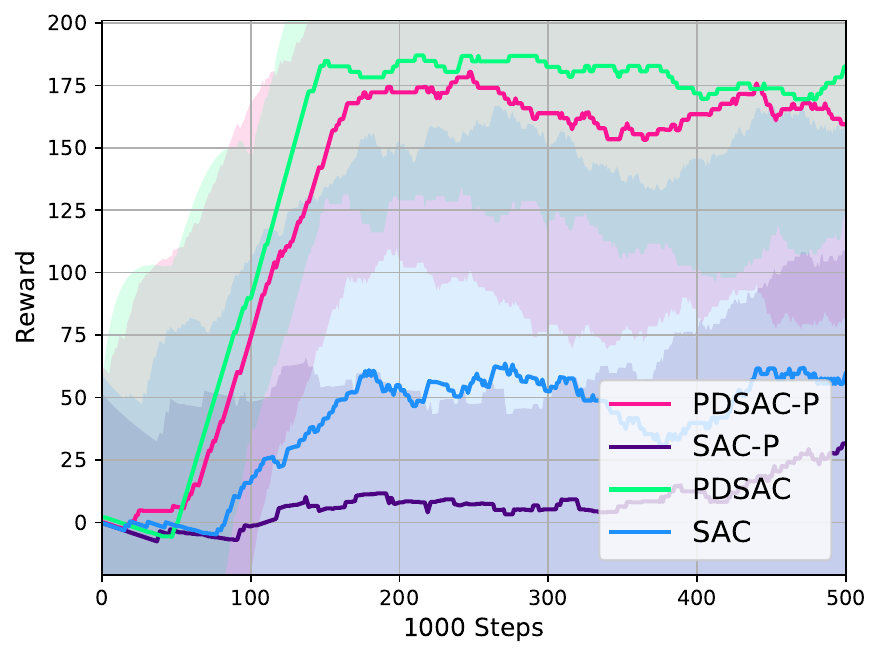}
         \caption{Environment 2.}
         \label{fig:stage2}
     \end{subfigure}
     \begin{subfigure}{0.325\textwidth}
         \includegraphics[width=\textwidth]{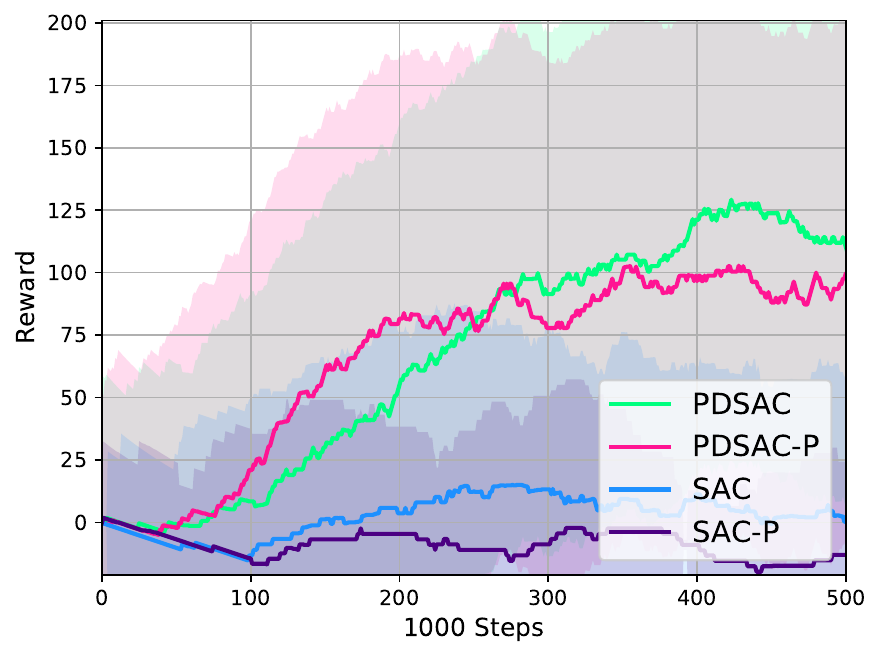}
         \caption{Environment 3.}
         \label{fig:stage3}
     \end{subfigure}
     \caption{Training rewards for all methods in the three environments.}
     \label{fig:stages}
\vspace{-5mm}
\end{figure*}

The training progress was demonstrated through reward graphs for each environment and the effectiveness of the methods showed through three-dimensional navigation trials. First, the rewards achieved by the evaluation agent were plotted to analyze the progress of the training (Fig.~\ref{fig:stage1}). The $y$-axis represents the moving average of rewards, while the $x$-axis represents the number of thousands of network update steps. The temporal metric used was the number of steps, as the agents' episodes varied due to parallelization and the training duration averaged 12 hours for non-prioritized methods and 48 hours for prioritized methods. The training and evaluation were done on an NVidia RTX 2080 GPU, an Intel I7-7700k CPU with 8 cores, 16 threads, and 32 GB of RAM.

The first environment, being simply an empty space, present few difficulties to the methods. We can observe in Fig.~\ref{fig:stage1}, that just the parallel method was able to converge in the defined training steps. The fact that SAC does not converge demonstrates a certain lack of ability by the agents to control their actions, especially the altitude. Analyzing the graph it can be seen that the methods using prioritized memory converged faster and showed greater stability.

The rewards for training in the second environment are represented in Figure \ref{fig:stage2}. In the second environment, obstacles were introduced in the form of four pillars, making it difficult for agents to navigate. Even so, it can be observed that our method converges, but this time with greater difficulty. In Figure \ref{fig:stage2} it can be seen that PDSAC presents great performance while navigating in this environment. On the contrary, both SAC and its prioritized form still lack the necessary learning to successfully reach the goals.

The third environment has a higher complexity than the previous two, having several obstacles in all dimensions blocking the passage of the drone in certain directions. Consequently, the results obtained by the agents were lower than the others, reaching only half of the total rewards achieved. Figure \ref{fig:stage3} portrays the rewards obtained during the training of methods in the third environment.

The results obtained presented in Figure \ref{fig:stage3}, by the agents were inferior to the others, reaching only half of the total possible rewards. These results are also due to the use of random targets, placing them in difficult-to-access positions. In this environment, it is also noticed that the methods have some difficulty in controlling altitude, a fact that is amplified by the lack of an altitude sensor. It also demonstrated a lot of difficulty in avoiding irregular obstacles, pointing out, even more, the need for sensors that go beyond two dimensions.

With the trained methods, their performances in carrying out navigation with obstacle avoidance to targets in predetermined fixed positions were evaluated. 100 trials were performed, 25 for each target. First, evaluate them in Environment 1, which has no obstacles. In Table \ref{table:amb1}, the percentage data of the success rate in the evaluations, the average rewards obtained and their standard deviation can be observed.

\begin{table}[bp]
\vspace{-3mm}
\centering
\setlength{\tabcolsep}{15pt}
\caption{Test results for environment 1.}
\begin{center}
\begin{tabular}{l | c c}
\toprule
\textbf{Method} & \textbf{Success rate (\%)} & \textbf{Average reward ($\mu$)}
\\ \midrule
\textbf{SAC} & $50.00 \%$ & $88.88 \pm 110.55$
\\ 
\textbf{SAC-P} & $50.00 \%$ & $88.88 \pm 110.55$
\\ 
\textbf{PDSAC} & $100.00 \%$ & $200.00 \pm 0.00$
\\ 
\textbf{PDSAC-P} & $100.00 \%$ & $200.00 \pm 0.00$
\\ \bottomrule
\end{tabular}
\label{table:amb1}
\end{center}
\end{table}

It can be noted that only the methods without parallelization were not able to obtain $100 \%$ of success. This occurred because during training these methods were unstable and did not reach the desired reward values.



Table \ref{table:amb2} presents the data obtained during the evaluation in the second environment. For this environment, with obstacles, it was found that only the PDSAC method with prioritized memory achieved all objectives. In the methods with random memory, the PDSAC obtained $81 \%$ of success. With 4 obstacles in this environment, the difficulty was higher when compared to the first one.

\begin{table}[bp]
\vspace{-5mm}
\centering
\setlength{\tabcolsep}{15pt}
\caption{Test results for environment 2.}
\begin{center}
\begin{tabular}{l | c c}
\toprule
\textbf{Method} & \textbf{Success rate (\%)} & \textbf{Average reward ($\mu$)}
\\ \midrule
\textbf{SAC} & $0.00 \%$ & $-10.10 \pm 10.05$
\\ 
\textbf{SAC-P} & $25.25 \%$ & $35.75 \pm 95.97$
\\ 
\textbf{PDSAC} & $81.81 \%$ & $160.00 \pm 85.28$
\\ 
\textbf{PDSAC-P} & $100.00 \%$ & $200.00 \pm 0.00$
\\ \bottomrule
\end{tabular}
\label{table:amb2}
\end{center}
\end{table}



The worst results were obtained in the third environment, as demonstrated in training. Table \ref{table:amb3} presents this data. The method that obtained the best results was the PDSAC with prioritized memory, again confirming the superiority of this memory model and the parallel methods. The disparity of results in this environment can be explained by its complexity, which ended up harming all methods similarly. In addition, it was noted that the methods began to memorize the best trajectories during training and repeated these during the evaluations. Thus, in this environment, a significantly better method was not obtained. The results obtained through the methods used in the third environment were inferior to the others. It is believed that this inferiority is attributed mainly to the complexity of the environment in question, which presents obstacles in all directions and at variable altitudes.

\begin{table}[tp]
\centering
\setlength{\tabcolsep}{15pt}
\caption{Test results for environment 3.}
\begin{center}
\begin{tabular}{l | c c}
\toprule
\textbf{Method} & \textbf{Success rate (\%)} & \textbf{Average reward ($\mu$)}
\\ \midrule
\textbf{SAC} & $0.00 \%$ & $-14.74 \pm 8.84$
\\ 
\textbf{SAC-P} & $0.00 \%$ & $-18.38 \pm 5.47$
\\ 
\textbf{PDSAC} & $50.50 \%$ & $101.01 \pm 100.50$
\\ 
\textbf{PDSAC-P} & $61.61 \%$ & $120.80 \pm 101.01$
\\ \bottomrule
\end{tabular}
\label{table:amb3}
\end{center}
\vspace{-5mm}
\end{table}



\begin{figure}[h]
      \begin{subfigure}{0.1175\textwidth}
          \includegraphics[width=\textwidth]{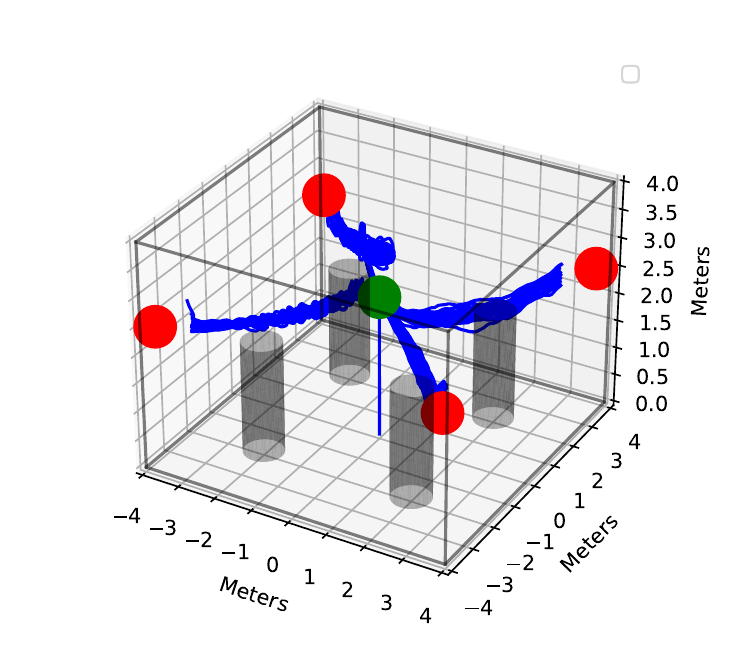}
          \caption{PDSAC-P.}
          \label{subfig:traj1_d4pg}
      \end{subfigure}
      \begin{subfigure}{0.1175\textwidth}
          \includegraphics[width=\textwidth]{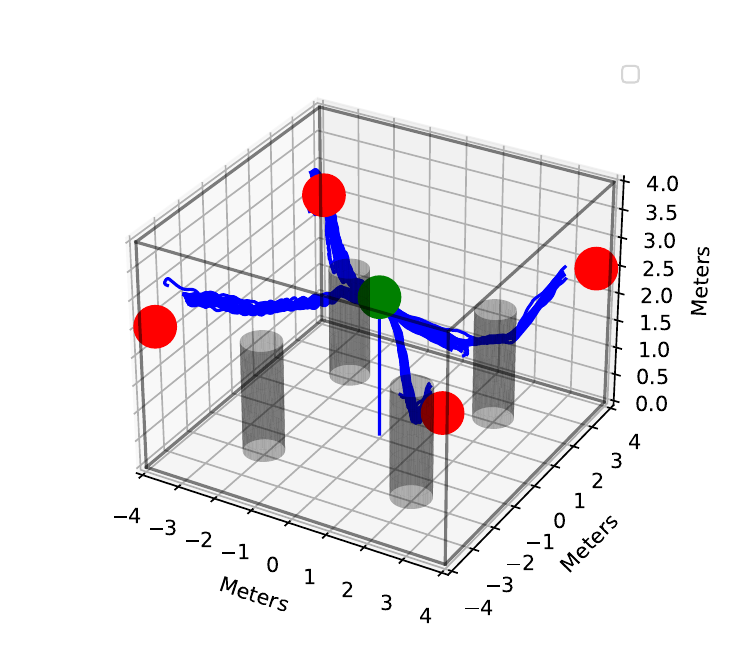}
          \caption{PDSAC.}
          \label{subfig:traj1_d4pgp}
      \end{subfigure}
      \begin{subfigure}{0.1175\textwidth}
          \includegraphics[width=\textwidth]{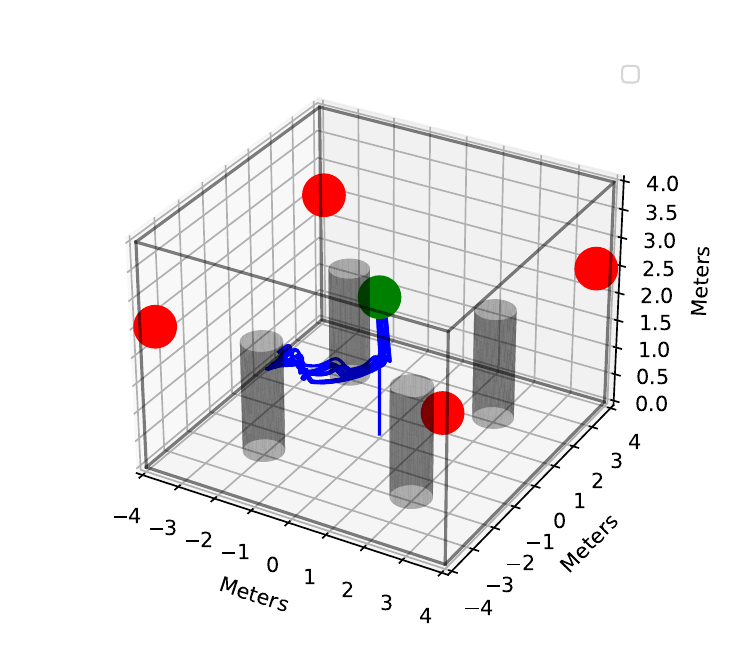}
          \caption{SAC-P.}
          \label{subfig:traj1_dsac}
      \end{subfigure}
      \begin{subfigure}{0.1175\textwidth}
          \includegraphics[width=\textwidth]{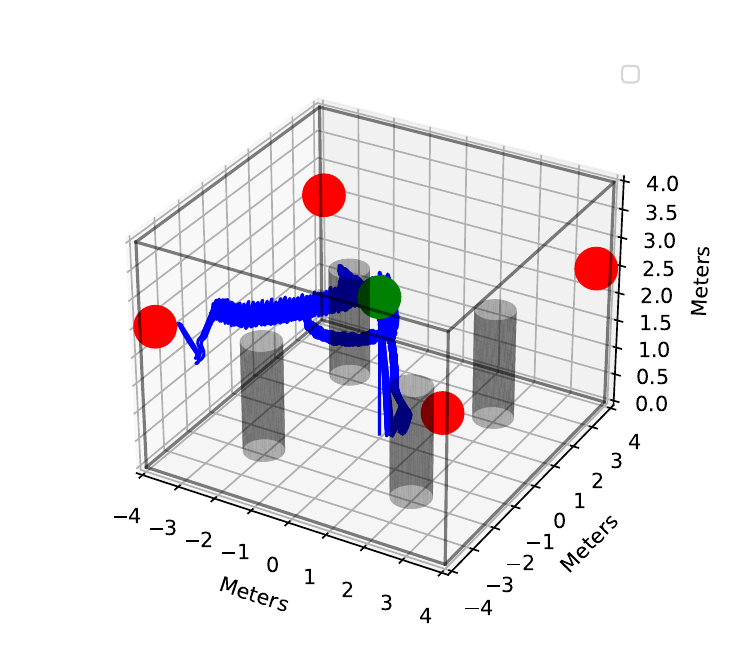}
          \caption{SAC.}
          \label{subfig:traj1_dsacp}
      \end{subfigure}
      \caption{Trajectories in environment 2. In blue we have the trajectories, green the start position, red the goals, and black the obstacles.}
      \label{fig:traj2}
 \end{figure}

To better assess the performance of the methods, we provide the trajectory executed on the evaluation of the second environment. As seen in Figure \ref{fig:traj2}, the targets are behind the obstacles, providing a greater range of decisions for agents. As the drone is capable of changing altitude, in this environment it is verified that the methods can fly over obstacles, a circumstance portrayed by the trajectories crossing obstacles. Analyzing the trajectories, it was concluded that the two versions of the PDSAC were able to fly over the obstacles, a fact that occurred due to the ability to gather greater amounts of data reducing the occurrence of catastrophic forgetting.

%% file: sections/8_conclusion.tex
\section{Conclusion}
\label{conclusion}



This study explored the use of DRL methods in UAV navigation, specifically introducing the PDSAC method that leverages parallelization and distributional capabilities. By employing multiple agents in parallel, training time was significantly reduced. Critical networks and prioritized memory were utilized to enhance performance, achieving the study's objectives and providing insights for further improvements. PDSAC outperformed previous methods, showcasing scalability, sample efficiency, and better representation of the value-action function. The fact that SAC had a bad performance exemplifies the occurrence of catastrophic forgetting. While with the parallel method, we have more data diversity ensuring that catastrophic forgetting is unlikely to occur. The incorporation of prioritized memory facilitated faster reward attainment during training in some cases, but more studies are necessary.


Future works include the addition of a vertical sensor for obstacle detection in building and simulating drones. Additionally, the application of these methods in real drones is also highlighted as a possible area for future research, thus proving the effectiveness of the methods in real scenarios.

%% file: sections/9_aknowledgement.tex
\section*{Aknowledgement}

The authors would like to thank the VersusAI team. This work was partly founded by the Technological University of Uruguay (UTEC), Federal University of Rio Grande (FURG), Federal University of Santa Maria (UFSM), CAPES, CNPq, and PRH-ANP.

%% file: sections/10_references.tex


\bibliographystyle{./bibliography/IEEEtran}
\bibliography{./bibliography/IEEEabrv,./bibliography/IEEEreferences}